\documentclass[letterpaper, 10 pt, conference]{ieeeconf}  

\IEEEoverridecommandlockouts                              

\usepackage[table]{xcolor}
\usepackage{graphicx}
\usepackage{caption}
\usepackage{amsmath}
\usepackage{amssymb}
\usepackage{amsfonts}
\usepackage{booktabs}
\usepackage{multirow}
\usepackage{tabularx}
\usepackage{xspace}
\usepackage{url}
\usepackage{bmpsize}
\usepackage{subcaption}
\usepackage{bm}
\usepackage{comment}
\usepackage{listings}
\usepackage{tcolorbox}

\newcommand{\Tref}[1]{Table~\ref{#1}}

\newcommand{\fref}[1]{Fig.~\ref{#1}}
\newcommand{\Fref}[1]{Figure~\ref{#1}}

\newcommand{\Sref}[1]{Section~\ref{1}}

\newcommand{\baselinename}{\emph{Name}}
\newcommand{\baselinefeature}{\emph{Feature}}
\newcommand{\baselineomni}{\emph{Omniglue}}

\graphicspath{{./figures_general_insertion/}}

\begin{document}
    \bstctlcite{IEEEexample:BSTcontrol}

    \title{\LARGE \bf
    Zero-Shot Peg Insertion: Identifying Mating Holes and\\ Estimating SE(2) Poses with Vision-Language Models}
    \author{Masaru Yajima$^{1,\ast}$, Kei Ota$^{2,\ast}$, Asako Kanezaki$^{1}$, Rei Kawakami$^{1}$
    \thanks{$^{1}$Masaru Yajima, Asako Kanezaki, and Rei Kawakami are with Institute of Science Tokyo, Tokyo, Japan.}
    \thanks{$^{2}$Kei Ota is with Mitsubishi Electric, Kanagawa, Japan. {\tt\small Ota.Kei@ds.MitsubishiElectric.co.jp}}
    \thanks{$^\ast$ indicates equal contribution.}
    }

    \maketitle
    \begin{abstract}
    Achieving zero-shot peg insertion, where inserting an arbitrary peg into an unseen hole without task-specific training, remains a fundamental challenge in robotics. This task demands a highly generalizable perception system capable of detecting potential holes, selecting the correct mating hole from multiple candidates, estimating its precise pose, and executing insertion despite uncertainties. While learning-based methods have been applied to peg insertion, they often fail to generalize beyond the specific peg-hole pairs encountered during training.
    Recent advancements in Vision-Language Models (VLMs) offer a promising alternative, leveraging large-scale datasets to enable robust generalization across diverse tasks. Inspired by their success, we introduce a novel zero-shot peg insertion framework that utilizes a VLM to identify mating holes and estimate their poses without prior knowledge of their geometry.
    Extensive experiments demonstrate that our method achieves 90.2\% accuracy, significantly outperforming baselines in identifying the correct mating hole across a wide range of previously unseen peg-hole pairs, including 3D-printed objects, toy puzzles, and industrial connectors. Furthermore, we validate the effectiveness of our approach in a real-world connector insertion task on a backpanel of a PC, where our system successfully detects holes, identifies the correct mating hole, estimates its pose, and completes the insertion with a success rate of 88.3\%.
    These results highlight the potential of VLM-driven zero-shot reasoning for enabling robust and generalizable robotic assembly.
\end{abstract}

    \section{Introduction}\label{sec:introduction}
Robotic assembly is a cornerstone of modern manufacturing, automating production lines and enhancing efficiency. While robots excel in structured pick-and-place tasks, complex assemblies---such as inserting an unknown electrical connector into a mating socket---require precise manipulation and object understanding. Among these tasks, peg-in-hole insertion is fundamental, forming the backbone of many component integration processes. Traditional peg-in-hole assembly relies on human-engineered heuristics, requiring manual adjustments or redesigns for each variation. However, with the increasing shift toward high-mix, low-volume production, robotic systems must adapt to assemble unknown mating parts without prior knowledge~\cite{wang2018thefuture}.

\begin{figure}[t]
    \centering
    \includegraphics[width=\linewidth]{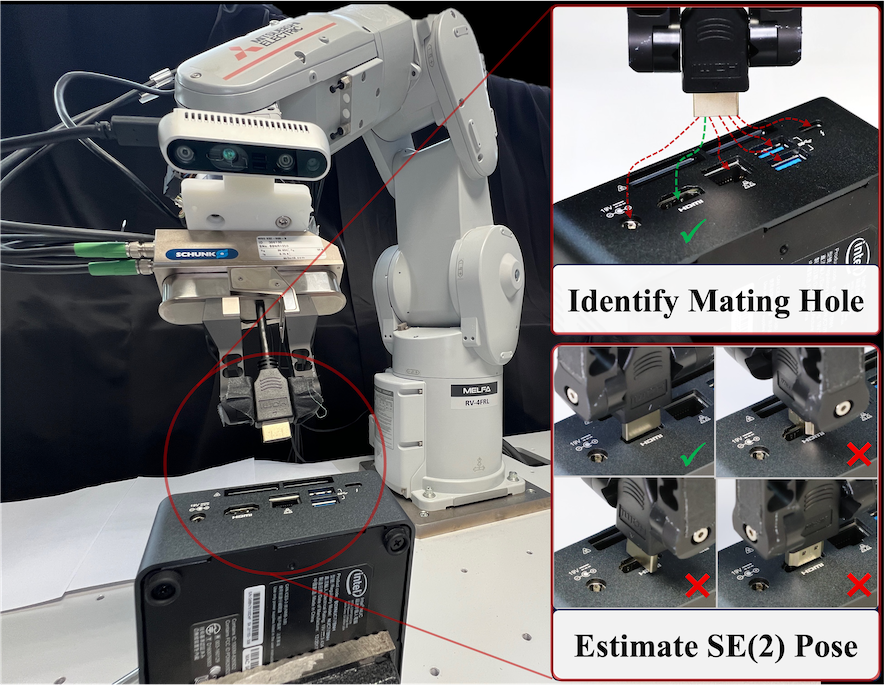}
    \caption{This work tackles the challenge of inserting an \emph{arbitrary} peg into an \emph{previously unseen} hole without prior knowledge of its type and geometry. We propose a novel framework that leverages a VLM to identify the correct mating hole (top right) and estimate its SE(2) pose (bottom right) in a zero-shot manner. Our method enables robust generalization across diverse peg-hole pairs, outperforming conventional approaches.}
    \vspace{-3mm}
    \label{fig:intro_fig}
\end{figure}

A common example of this challenge is a robot attempting to insert an HDMI connector into a back panel of a PC, as shown in \fref{fig:intro_fig}. 
This process consists of two main phases~\cite{xu2019compare,jiang2022areview}: \emph{search} and \emph{insertion}. During the \emph{search} phase, the robot determines which holes are compatible with the given peg and identifies the precise pose of the corresponding hole. The \emph{insertion} phase then focuses on generating assembly actions to guide the grasped peg into the target hole.

While significant research has focused on improving the \emph{insertion} phase through predefined search paths~\cite{chhatpar2001search,jiang2022the}, learning-based methods~\cite{zhang2023vision}, and reinforcement learning~\cite{schoettler2020deep}, the equally critical \emph{search} phase remains largely underexplored.
Current approaches inherently assume prior knowledge of the correct mating part and its approximate location, thereby bypassing the fundamental challenge of identifying the correct hole from multiple candidates. 
Prior work~\cite{pauli2001vision,stemmer2007ananalytical,song2017guidance} has explored peg-hole identification, but is largely restricted to primitive geometric shapes like circles and rectangles,  failing to generalize to complete shapes such as industrial connectors, which often have complex contours, keying features. Additionally, estimating the relative pose of an \emph{unseen} hole with respect to \emph{unseen} peg remains an unsolved challenge.
This work bridges this critical gap by introducing a zero-shot method for peg-hole identification and pose estimation. Given a set of candidate holes, our framework identifies the optimal mating hole for the grasped peg and accurately estimates its SE(2) pose---without requiring task-specific training. Addressing this overlooked problem is crucial for advancing autonomous assembly in industrial automation, robotics, and manufacturing.

To develop a method that generalizes across a diverse range of peg-hole pairs---from simple geometric shapes to complex industrial connectors such as HDMI---we leverage vision-language models (VLMs), which are pre-trained on vast image-text datasets and excel in visual recognition without fine-tuning~\cite{zhang2024vision}.
In this work, we harness VLMs for zero-shot peg insertion. Given images of a peg and holes, we prompt the VLM to determine their compatibility, selecting the best mating hole from multiple candidates. We then estimate its relative pose by prompting the VLM with different insertion angles.
To maximize effectiveness, we refine input image selection, prompt design, and output processing. Specifically, we incorporate two viewpoints per peg-hole pair and constrain the VLM's response to a binary \texttt{Yes} or \texttt{No}, extracting confidence scores to resolve ambiguity when multiple holes receive positive responses.

Extensive experiments validate our approach across diverse peg-hole pairs, including industrial connectors, toy puzzles, and 3D-printed parts, achieving a 90.2\% success rate, outperforming existing methods. Furthermore, we conduct ablation studies that systematically modify the input and output of the VLM to validate the effectiveness of our approach.
Moreover, we evaluate our method in a more realistic scenario, inserting industrial connectors into the back panel of a PC, achieving a 88.3\% success rate. In summary, our contributions are as follows:
\begin{enumerate}
    \item We introduce the first VLM-based zero-shot method for identifying mating holes and estimating SE(2) poses in arbitrary peg-hole pairs, eliminating the need for task-specific training.
    \item We conduct extensive real-world evaluations across diverse peg-hole pairs, demonstrating our method's robustness in 3D-printed parts, toy puzzles, and industrial connectors.
    \item We integrate the proposed perception system into a full insertion pipeline and evaluate its performance in real-world insertion experiments.
\end{enumerate}
To the best of our knowledge, this is the first application of VLMs for zero-shot general peg insertion, paving the way for more adaptable and autonomous robotic assembly.

    \section{Related Work}\label{sec:related_work}

\textbf{Learning-based Methods for Insertion.}
Reinforcement learning (RL) has emerged as a promising approach for peg-in-hole insertion, generating effective control commands for precise assembly~\cite{schoettler2020deep}. However, these methods often suffer from overfitting to specific training objects, limiting their real-world applicability. Strategies such as offline meta-RL pretraining~\cite{zhao2022offline} and real-world fine-tuning~\cite{nair2023learningonthejob} have attempted to improve generalization, but they still require real-world training data and struggle with diverse object categories. Notably, RL models trained on 3D-printed objects frequently fail when applied to industrial connectors---a critical limitation that our work directly addresses by circumventing the need for task-specific training.

\textbf{Shape Correspondence Estimation.}
Object correspondence estimation has been widely studied in computer vision, primarily using hand-crafted features or learning-based methods for RGB images~\cite{florence2018dense,jiang2024Omniglue,li2024matching} and point clouds~\cite{besl1992icp,zeng20173dmatch,zhong2023chsel}. While these methods excel at aligning geometrically similar points, they struggle to identify mating parts, which require reasoning beyond shape similarity. Even state-of-the-art approaches such as OmniGlue~\cite{jiang2024Omniglue} fail in our experiments, highlighting a fundamental gap that our method addresses.
Despite its importance, visual identification of mating parts remains largely unexplored. Traditional feature-matching techniques for peg-hole alignment~\cite{pauli2001vision,stemmer2007ananalytical,song2017guidance} fail with intricate geometries, while learning-based models like Form2Fit~\cite{zakka2020form2fit} rely on extensive labeled datasets yet struggle with complex shapes such as DisplayPort connectors. Tactile-based approaches~\cite{ota2023tactile} offer an alternative but require hundreds of physical interactions, making them impractical. In contrast, our method leverages VLMs' strong generalization capabilities to accurately identify mating parts from images, enabling robust and scalable correspondence estimation without task-specific training or costly data collection.



\textbf{Pose Estimation for Insertion.}
Accurate pose estimation of the target hole is crucial for successful peg insertion, enabling direct insertion or a more constrained search space. Existing methods rely on predefined object categories~\cite{gao2021kpam2}, shape priors from 3D CAD models~\cite{litvak2019learning}, or extensive task-specific training~\cite{zhang2023vision}. These constraints hinder generalization to unseen objects. In contrast, our approach eliminates these dependencies, enabling robust pose estimation across arbitrary geometries without requiring task-specific training.

\textbf{VLMs for Robots.}
The integration of VLMs into robotics has traditionally focused on linking pre-trained language models with robotic actions~\cite{jang2022bcz,shridhar2023perceiver,singh2023progprompt}. While these approaches have shown promise, they struggle with complex semantic reasoning and fail to generalize in open-world environments. More recent efforts leverage VLMs for task planning~\cite{saycan2022arxiv,liang2023code}, robotic manipulation~\cite{black2024pi_0}, and reward formulation~\cite{wang2024rlvlmf}, demonstrating their strong zero-shot generalization capabilities.
Building on this trend, our study introduces a novel application of VLMs for peg-in-hole assembly. Unlike prior work, we leverage a pre-trained VLM to both identify the correct mating part from a set of candidates and estimate its relative pose without requiring task-specific training. This enables zero-shot generalization across diverse peg-hole pairs, including industrial connectors, toy puzzles, and 3D-printed components, marking a significant step toward more adaptable robotic assembly systems.
Despite their strong generalization capabilities, VLMs can be sensitive to prompt design and may struggle with fine-grained geometric reasoning. To address this, we refine our approach by incorporating multiple viewpoints per peg-hole pair, designing structured prompts, and constraining VLM outputs to binary confidence-based responses. These strategies mitigate potential ambiguities and improve decision robustness.

    \section{Problem Statement}
This work addresses a fundamental and long-standing challenge in robotics: inserting an \emph{arbitrary} peg into an \emph{unseen} hole without task-specific training. 
Solving this requires a system that can detect potential holes, identify correct peg-hole pairs, estimate their precise poses, and perform insertion---all without prior knowledge of the specific task (see \fref{fig:intro_fig}).
Despite its apparent simplicity, peg-in-hole insertion remains a critical benchmark for robotic manipulation. Given the complexity of solving this problem in its entirety, we introduce reasonable assumptions to make progress within a well-defined scope.

\textbf{Task Assumptions.}
To primarily evaluate the perception capability and robustness of our system on unseen peg-hole pairs, we define the following simplifying assumptions:

\begin{enumerate}
    \item The pose of the grasped peg is known.
    \item The rough positions of holes are known.
    \item The surface for the holes is perfectly leveled.
\end{enumerate}
Assumption (1) can be relaxed by integrating a vision-based or visuo-tactile system for real-time in-hand pose tracking~\cite{ota2024autonomous}. Assumption (2) removes the need for global object search, allowing focus on peg-hole matching and pose estimation. Assumption (3), commonly used in perception-focused insertion tasks~\cite{triyonoputro2019quickly,haugaard2021fast}, can be relaxed by estimating and compensating for surface tilt~\cite{zhang2023vision}.



    \section{Method}\label{sec:system}
In this section, we introduce a robust method that leverages VLMs to achieve generalized zero-shot peg insertion. Our framework consists of four key components: (1) utilizing a VLM to assess peg-hole compatibility and identify the most suitable match, (2) employing a VLM to estimate the SE(2) pose, (3) detecting and filtering candidate holes using an instance segmentation model, and (4) integrating a stiffness controller to ensure stable and adaptive insertion. 

\subsection{Peg-Hole Matching using VLMs} \label{subsec:matching}
To achieve zero-shot peg-hole matching across diverse objects without requiring additional training, we leverage a VLM to identify the most compatible peg-hole pair. Effectively applying a general-purpose VLM to this specialized task requires a carefully structured approach to input design, prompt formulation, and output interpretation.
To accomplish this, we strategically present the VLM with peg and hole image pairs and construct a task-specific prompt that directs the model to evaluate their compatibility. The effectiveness of this approach relies on three critical factors: the selection of input images, the design of an effective prompt, and the interpretation of the VLM's responses, detailed as follows.

\textbf{Input Images.}
To accurately determine the best-matching hole for a given peg, we capture multi-view images from different angles for both the peg and hole. These images are denoted as $I^\text{p}_{n}, I^\text{h}_{m,n}$, where $I^\text{p}$ and $I^\text{h}$ represent images of the peg and hole, respectively.
The index $m\in \{ 1, ..., M\}$ refers to the $m$-th candidate hole identified through the detection process in Section~\ref{subsec:hole}. Similarly, $n \in \{ 1, ..., N \}$ represents different viewpoints captured by the robot. In this study, we use $N=2$, capturing images from a top-down perspective and a $30$-degree angled view to ensure comprehensive shape analysis.
Incorporating multi-view images significantly enhances the VLM's ability to assess peg-hole compatibility by providing richer spatial context and reducing ambiguity in shape analysis. Our ablation study confirms that using two viewpoints substantially improves accuracy over a single-view approach, reinforcing the importance of multi-angle image acquisition in our framework.

\textbf{Prompt.}
One of the primary challenges of applying a general-purpose VLM to peg-hole matching is ensuring that the model accurately interprets the task within its broad knowledge domain. Since VLMs are designed for diverse and open-ended applications, a well-crafted prompt is essential to focus the model's attention on the relevant spatial and geometric relationships while eliciting reliable outputs.
To achieve this, we design a structured prompt that explicitly defines the relationship between the peg and hole images:
\begin{tcolorbox}[colframe=black!50, colback=gray!5, coltitle=black, title=Prompt to VLM, fonttitle=\bfseries, fontupper=\small]
    \texttt{<image1> This is a cross-sectional image of a peg.} \\
    \texttt{<image2> This is another image of a peg from a different angle.} \\
    \texttt{<image3> This is a cross-sectional image of a hole.} \\
    \texttt{<image4> This is another image of a hole from a different angle.} \\
    \texttt{Can the peg in images 1 and 2 be perfectly inserted into the hole in images 3 and 4?} \\
    \texttt{Please answer with only yes or no.}
\end{tcolorbox}
Each \texttt{<image>} token in the prompt corresponds to an input image in the following order: $I^\text{p}_{1}$, $I^\text{p}_{2}$, $I^\text{h}_{m,1}$, and $I^\text{h}_{m,2}$.
By framing the problem as a binary classification task, this structured prompt minimizes ambiguity and ensures that the VLM focuses on functional compatibility rather than general object recognition. This design enhances scalability and generalizability, making it applicable to various object shapes and geometries without requiring task-specific retraining.

\textbf{Interpreting VLM Responses.} \label{subsec:output_processing}
Given the input images and structured prompt, the VLM produces a binary response either \texttt{Yes} or \texttt{No}, indicating whether the peg can be successfully inserted into the hole. However, two key challenges arise: (1) multiple candidate holes may receive a \texttt{Yes} prediction for a single peg, leading to ambiguity in selection, and (2) the VLM may predict \texttt{No} for all holes, resulting in no valid match. 
To overcome these challenges, we extract and analyze the generation probabilities associated with the VLM's responses and incorporate them into a confidence-based ranking strategy for robust decision-making.

\begin{figure}[t]
    \centering
    \includegraphics[width=0.9\linewidth]{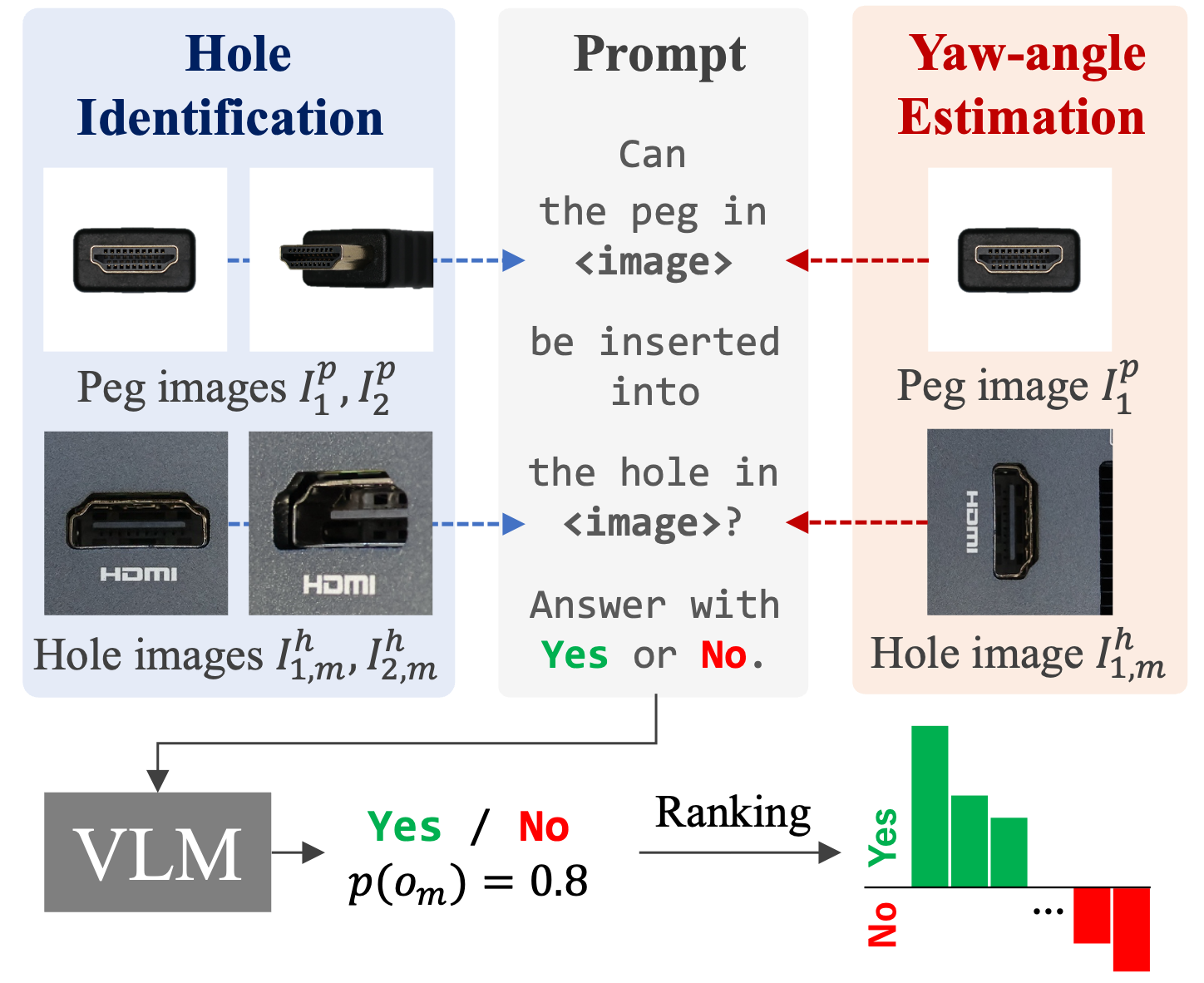}
    \caption{
    We provide the VLM with the peg image(s) and the candidate hole image(s) along with a prompt. The VLM determines whether the given peg and hole constitute a valid match by outputting either \texttt{Yes} or \texttt{No}, accompanied by the corresponding generation probability $p(o_m)$. This process is repeated for all candidate holes and the generated probabilities are used to rank the most suitable candidate, enabling the identification of the most compatible hole.}
    \vspace{-5mm}
    \label{fig:pipeline}
\end{figure}

Let the output of the $m$-th hole be $o_m \in \{\texttt{Yes}, \texttt{No}\}$, with its corresponding probability $p(o_m)$, derived from the final-layer softmax logits of the VLM~\cite{li2025llavaonevision}.
To ensure an optimal selection process, we implement the following ranking strategy: Holes predicted as \texttt{Yes} are ranked in descending order of confidence, prioritizing high-confidence matches. Holes predicted as \texttt{No} are ranked in ascending order, ensuring that less confident rejections are reconsidered if no valid matches exist. Finally, the highest-ranked hole is selected as the most probable match (see Fig.~\ref{fig:pipeline}).
This probability-aware ranking mechanism ensures a more refined and reliable peg-hole matching process by leveraging not just binary outputs, but also the model's internal confidence levels to resolve ambiguities and maximize the likelihood of correct pairings.

\subsection{SE(2) Pose Estimation using VLMs}
Once a peg-hole pair is identified, precise pose estimation is essential for successful insertion. Among the pose parameters, yaw-angle estimation is particularly critical, as localization errors can be compensated with some search approaches, such as spiral search as we will show in our experiments. 
Traditional methods~\cite{pauli2001vision,stemmer2007ananalytical,song2017guidance} rely on contour-based features but struggle to distinguish between visually similar yet functionally incompatible orientations. A clear example is the USB-A connector, whose contour appears identical when flipped 180 degrees but fails to insert due to internal misalignment. Contour-based methods cannot capture these fine-grained rotational differences, making them unreliable for precise peg-hole alignment.

To overcome these limitations, we propose a VLM-based approach that reasons beyond surface-level appearance, leveraging deep semantic understanding to infer spatial relationships critical for functional alignment. Our method decomposes SE(2) pose estimation into two key stages: (1) yaw-angle estimation, where a VLM-based classification approach determines the precise hole orientation, and (2) localization, where the hole's centroid is extracted from the segmented pointcloud for accurate positional alignment.

\textbf{Yaw-Angle Estimation via VLM.}
Since most pegs and holes have flat edges, aligning these edges with the image axes provides a reliable reference for orientation estimation. To determine the relative yaw-angle difference between the grasped peg and the target hole, we formulate the problem as a four-class classification task. 

\begin{figure}
    \centering
    \begin{subfigure}{0.24\linewidth}
        \includegraphics[width=\linewidth]{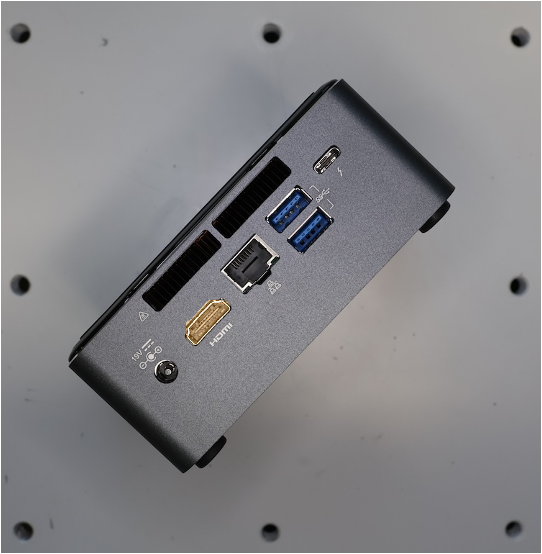}
        \subcaption[]{}
        \label{fig:yawangle_all}
    \end{subfigure}
    \begin{subfigure}{0.24\linewidth}
        \includegraphics[width=\linewidth]{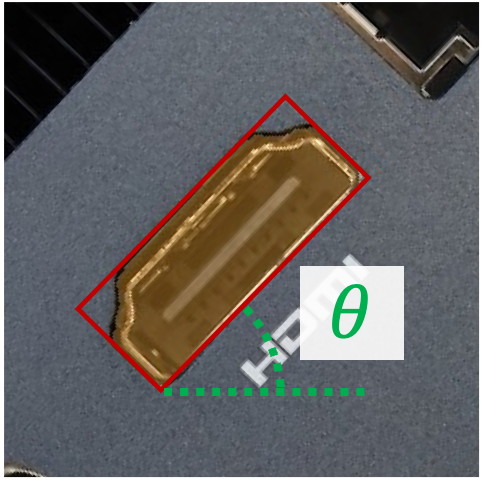}
        \subcaption[]{}
        \label{fig:yawangle_tilted}
    \end{subfigure}
    \begin{subfigure}{0.24\linewidth}
        \includegraphics[width=\linewidth]{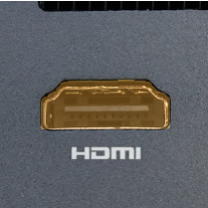}
        \subcaption[]{}
        \label{fig:yawangle_canonical}
    \end{subfigure}
    \begin{subfigure}{0.24\linewidth}
        \includegraphics[width=\linewidth]{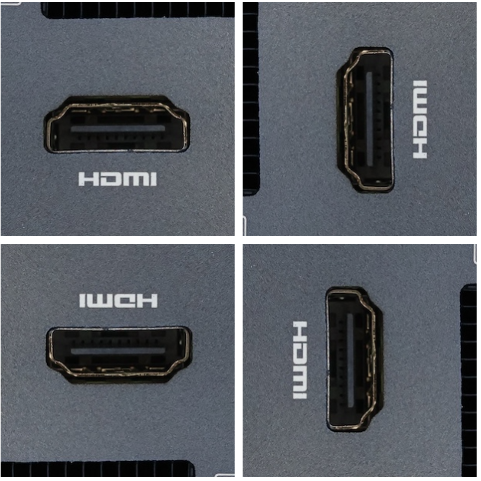}
        \subcaption[]{}
        \label{fig:yawangle_four}
    \end{subfigure}
    \caption{Yaw angle estimation process: (a) input RGB image $I^h_\text{all}$; (b) minimum bounding rectangle obtained from the segmentation mask; (c) image rotated to align the rectangle with the camera axis; (d) images rotated at 0$^{\circ}$, 90$^{\circ}$, 180$^{\circ}$, and 270$^{\circ}$ from the generated image at step (c), which are inputted to the VLM (see \fref{fig:pipeline}). }
    \vspace{-5mm}
    \label{fig:yawangle}
\end{figure}

To achieve this, we first extract the minimum bounding rectangle from the hole image $I_{m,1}^\textrm{h}$ using its associated segmentation mask obtained from the detection process in Sec.~\ref{subsec:hole}, as shown in \fref{fig:yawangle_tilted}. Defining the \emph{canonical image} as an image whose bounding rectangle aligns with the image axes, we then compute a rotation angle $\theta \in [0, 90)$ from the $x,y$ coordinates consisting of the bounding box. 
We apply a rotation matrix that rotates the image by $-\theta$ along the $z$-axis (camera's viewing axis) to align the bounding box with the image axes, transforming the original image into a canonical image as shown in \fref{fig:yawangle_canonical}.

Once the canonical orientation is established, we generate three additional images by rotating the canonical image by $90^\circ, 180^\circ,$ and $270^\circ$ (\fref{fig:yawangle_four}). These rotations allow the VLM to compare the peg image $I^\textrm{p}_{1}$ against all possible orientations, ensuring that the best-matching alignment is identified. 

Finally, we input the peg image and the generated four axis-aligned images into the VLM using the same prompt as peg-hole matching in Sec.~\ref{subsec:matching} and apply the same post-processing method to select the best alignment. Once we identify the best matching image, we determine the yaw-angle orientation as $\theta^\text{yaw}=\theta^\text{rotate}+\theta$, where $\theta^\text{rotate}\in \{0, 90, 180, 270 \}$ represents the applied rotation angle.

\textbf{Localization.}
The hole's position is determined as the centroid of the segmented pointcloud in the world frame, denoted as $\hat{\mathbf{p}}_m^h$, and computed using Eq.~\eqref{eq:position} in Sec.~\ref{subsec:matching}.

\subsection{Detection of Candidate Holes}\label{subsec:hole}
Efficient and accurate detection of candidate holes is the foundation of our zero-shot peg insertion pipeline. To achieve this, we employ Fast-SAM~\cite{zhao2023fast}, a state-of-the-art general-purpose segmentation model, to extract potential hole locations from an image captured 0.5 m above the table, ensuring comprehensive coverage of the object's surface.

Since Fast-SAM generates numerous segmentation masks across the entire image, many of these do not correspond to actual holes, often representing background objects or irrelevant structures. To eliminate such false positives, we retain only masks with heights greater than a predefined threshold, $z_\text{threshold}$, measured from the table.

To determine whether the $k$-th segmented image $I^\text{h}_{k,\text{seg}}$ satisfies this criterion, we first compute its corresponding segmented point cloud in the world frame, $P^\text{h}_k$.\footnote{Unless otherwise specified, all terms are expressed in the world frame $w$, with the origin at the center of the robot base.} This is done by projecting the segmentation mask onto the depth image to extract points in the camera frame, $^cP^\text{h}_k$. These points are then transformed into the world frame using the gripper pose ($^wT^g$) obtained through forward kinematics, along with the wrist-mounted camera pose ($^gT^c$) obtained through calibration before experiments.
We then compute the average position of the points as follows.
\begin{equation}
\bar{\mathbf{p}}^\text{h}_k = \begin{bmatrix} \bar{x}_k^\text{h} \\ \bar{y}_k^\text{h} \\ \bar{z}_k^\text{h} \end{bmatrix} = \frac{1}{| P^\text{h}_k |} \sum_{\mathbf{p} \in P^\text{h}_k} \mathbf{p}. \label{eq:position}
\end{equation}
We will eliminate $k$-th segment if $\bar{z}_k^\text{h} < z_\text{threshold}$. We specifically set $z_\text{threshold}=50$ mm in our experiments.

\subsection{Spiral Search-based Insertion}
Once the correct peg-hole pair and precise pose are determined, the final step is physically inserting the peg. Ideally, the robot should insert the peg directly using a pick-and-place motion. However, pose estimation errors may arise due to imperfect segmentation masks, extrinsic camera calibration inaccuracies, or yaw-angle uncertainty in the VLM's predictions. These errors can lead to misalignment, preventing successful insertion. To compensate, we employ a spiral search strategy, enabling the robot to refine the peg's alignment through local exploration.

We first command the robot to move towards the estimated hole position $\hat{\mathbf{p}}_m^\text{h}$ with a $3$ mm above position.
We then activate the robot's default stiffness controller~\cite{salisbury1980activestiffness} to establish a gentle contact, gradually lowering the peg until the measured $z$-axis contact force reaches a predefined threshold of 1 N.
Once contact is established, the robot initiates a spiral search, incrementally adjusting its position to explore potential misalignments. Thanks to the flat surface of the object and the compliance of the stiffness controller, the peg automatically inserts itself into the hole once proper alignment is achieved.

    \section{Experiments}\label{sec:results}
In this section, we rigorously evaluate our method on a diverse set of peg insertion tasks performed on a real system, demonstrating its effectiveness in real-world scenarios. First, we assess the part-mating performance using VLMs, highlighting their ability to generalize across unseen peg-hole pairs. We then conduct a comprehensive ablation study to analyze the impact of different VLM usage strategies on system performance. Finally, we validate the full pipeline, encompassing detection, identification, pose estimation, and insertion, to showcase the system's end-to-end robustness. Importantly, all test objects are novel, and the model operates without any trainable modules, underscoring its zero-shot generalization capability.
We use the following system for the experiments.

\textbf{Robot Platform.}\,
The MELFA RV-4FRL, a high-precision 6-DoF industrial robot, is used in this study. This robot is equipped with a Force-Torque sensor mounted on its wrist for allowing the stiffness-controlled approach.

\textbf{Vision Sensor.}\,
We mount an Intel RealSense D435 camera on the robot's wrist for capturing RGB-D data with a resolution of $(1280, 720)$ pixels. This setup allows the system to detect and localize target holes corresponding to given pegs. 
We calibrate the camera's extrinsic parameters to the wrist pose before conducting experiments (see \fref{fig:intro_fig}).

\subsection{Peg-Hole Matching Performance} \label{subsec:exp_perception}

\begin{figure}[t]
    \centering
    \begin{subfigure}{0.578\linewidth}
        \includegraphics[width=\linewidth]{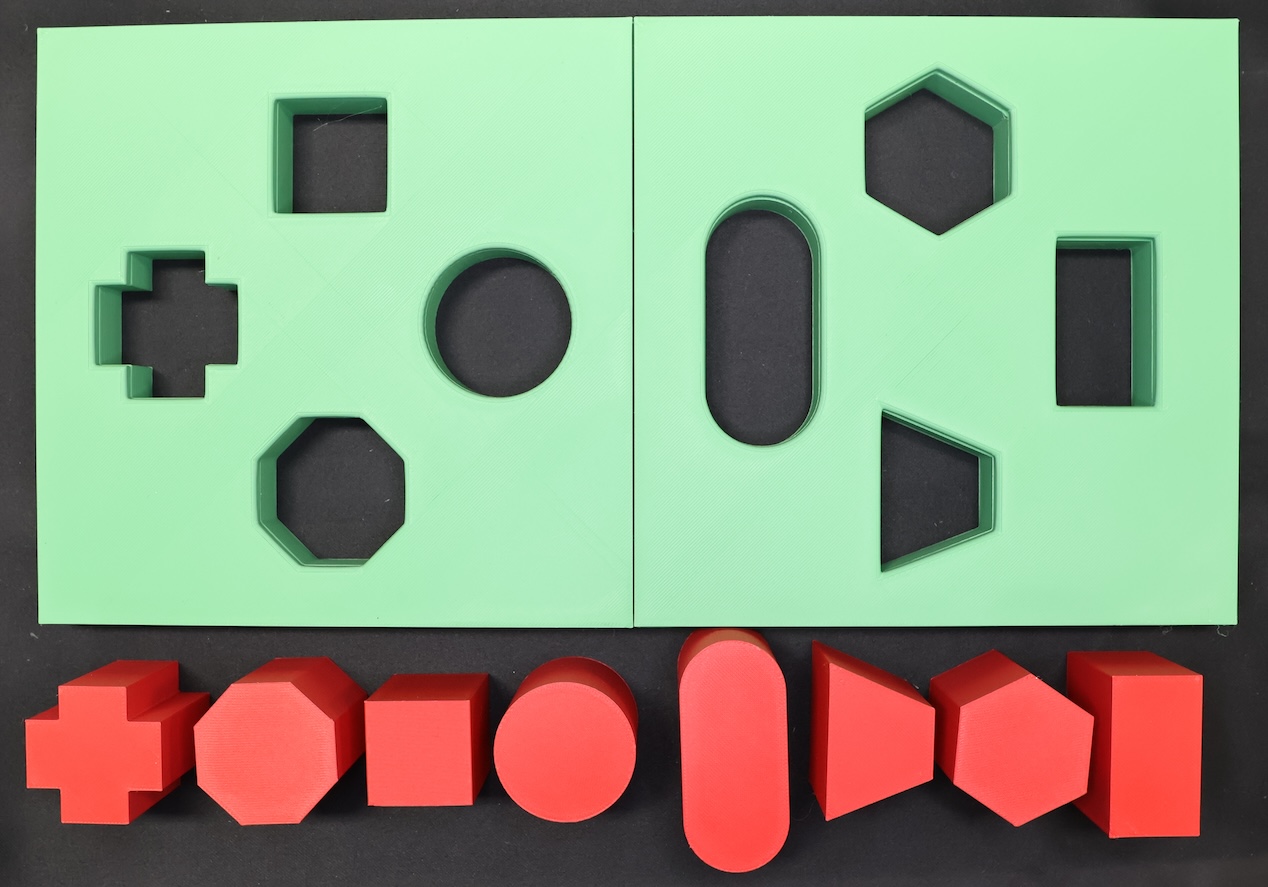}%
        \caption{3D-printed parts}
    \end{subfigure}
    \begin{subfigure}{0.405\linewidth}
        \includegraphics[width=\linewidth]{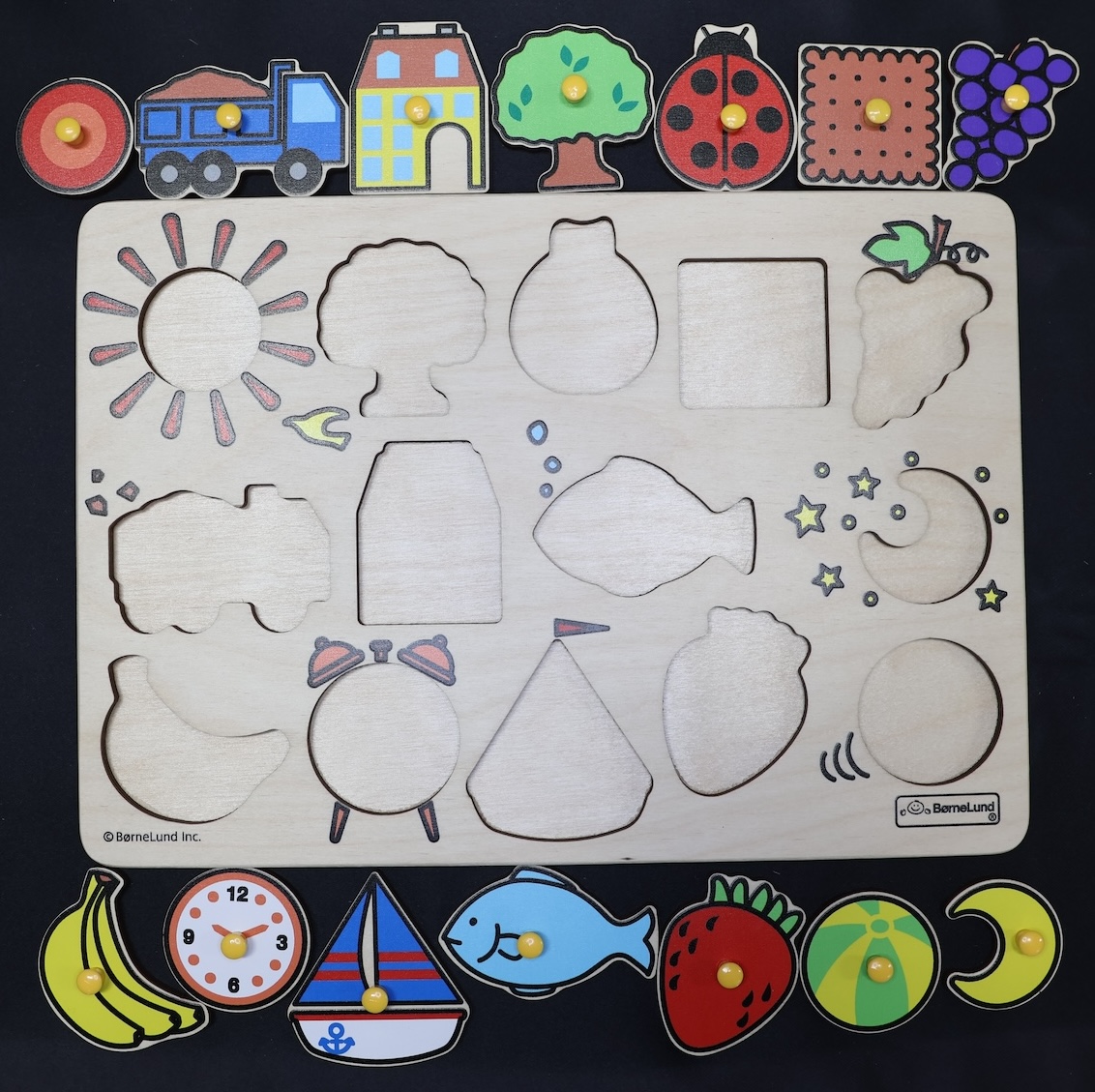}%
        \caption{Toy puzzles}
    \end{subfigure}
    \begin{subfigure}{\linewidth}
        \includegraphics[width=\linewidth]{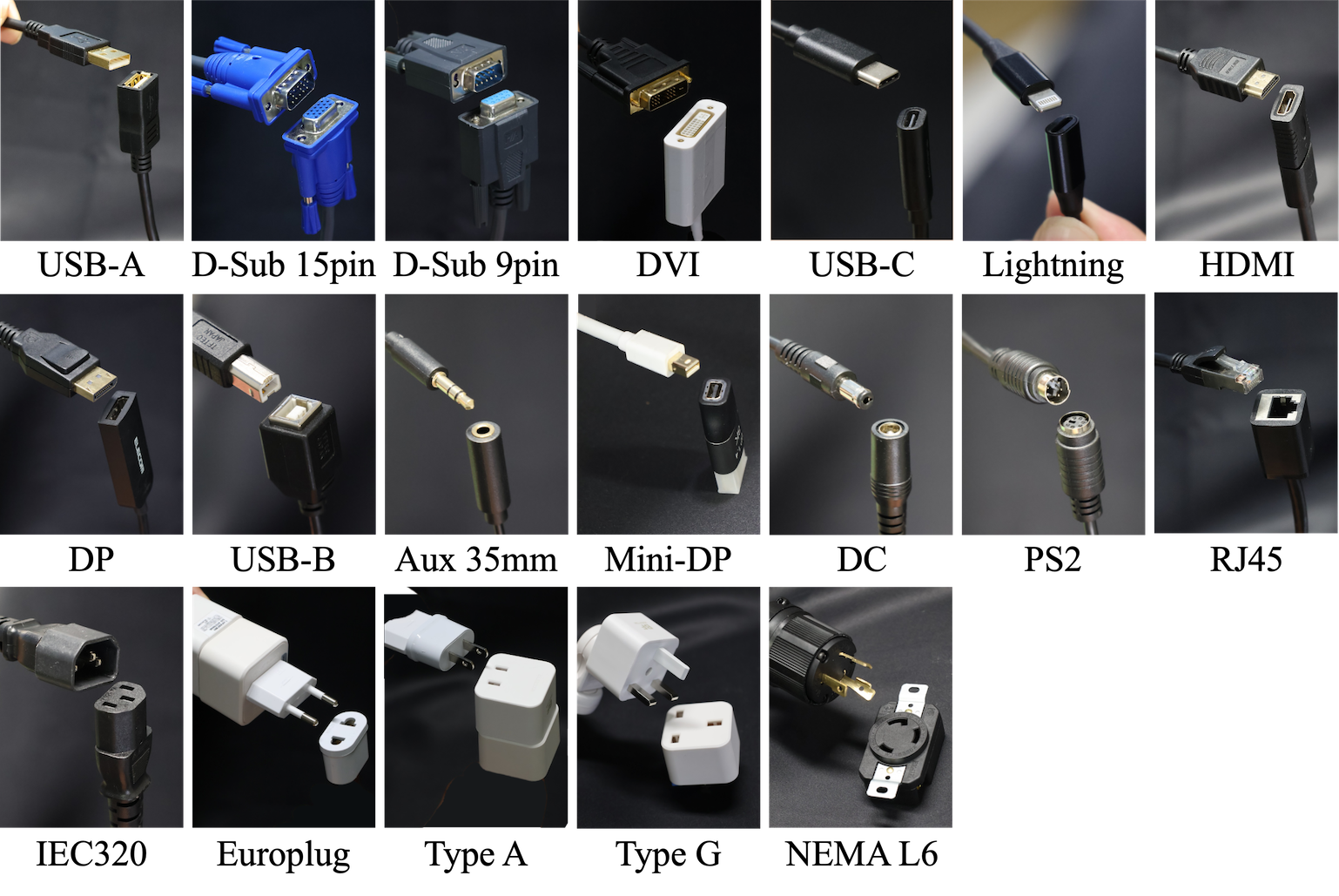}%
        \vspace{-2mm}
        \caption{Industrial connectors}
    \end{subfigure}
    \caption{Diverse peg-hole pairs used for our experiments.}
    \vspace{-3mm}
    \label{fig:objects}
\end{figure}

\textbf{Settings.}
We use three different sets of peg-hole pairs as shown in \fref{fig:objects}: 3D-printed parts with 8 pairs, toy puzzles with 14 pairs, and industrial connectors with 19 pairs. For each peg-hole pair, we obtain cross-sectional and angled images by sending commands to the robot to move to the predefined positions (see \fref{fig:pipeline}).

We use LLaVA-Onevision~\cite{li2025llavaonevision} as the VLM for peg-hole matching.

\textbf{Baselines.}
We compare our method with two baselines. The first baseline, \baselinefeature, extracts the class token feature vector from the output of the VLM's image encoder and performs the matching by calculating the cosine similarity between the feature vectors.
Second, to compare our method with the edge-based similarity matching approach used in prior work~\cite{pauli2001vision,stemmer2007ananalytical,song2017guidance}, we evaluate our method against \baselineomni, where we extract edges using EDTER~\cite{Pu2022EDTER} and perform matching using Omniglue~\cite{jiang2024Omniglue}, a state-of-the-art feature matching method widely used for identifying correspondence points between arbitrary image pairs.

\textbf{Metric.}
The performance is assessed by classification accuracy. In the \baselineomni\ baseline, the hole with the most detected feature points is chosen as the match.

\begin{table}[t]
    \centering
    \caption{Success rate. The numbers with the yellow background show the best results.
    }
    \begin{tabular}{c|ccc} \toprule
         & 3D print & Toy puzzle & Connectors \\ \midrule
        \baselinefeature & 1/8 & 4/14 & 7/19 \\
        \baselineomni~\cite{jiang2024Omniglue} & 4/8 & 0/14 & 4/19\\
        Ours & \cellcolor{yellow!50}7/8 & \cellcolor{yellow!50}14/14 & \cellcolor{yellow!50}16/19 \\ \bottomrule
    \end{tabular}
    \vspace{-3mm}
    \label{tab:perception_results}
\end{table}

\begin{figure}[t]
    \centering
    \includegraphics[width=0.9\linewidth]{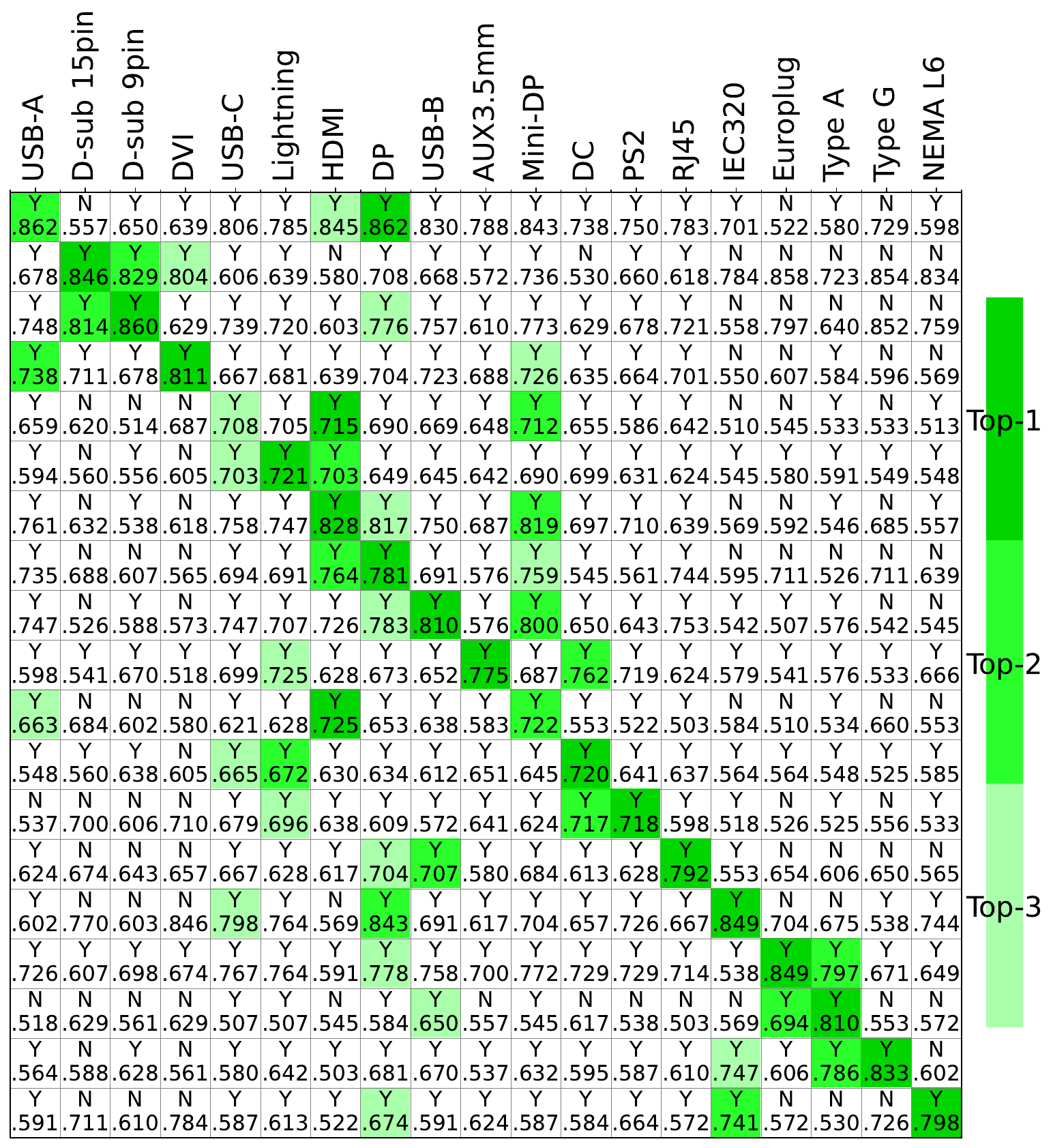}
    \caption{Responses (\emph{Yes} or \emph{No} shown as \texttt{Y} or \texttt{N}) and associated probabilities for each peg-hole pair, generated by our method for industrial connectors. The three shades of green represent top-1, top-2, and top-3. The table shows that top-3 accuracy is $100\%$, so the robot can insert the peg into the mating hole within $3$ trials if we allow multiple trials.}
    \vspace{-5mm}
    \label{fig:confusion_matrix}
\end{figure}

\textbf{Results.}
\Tref{tab:perception_results} presents the success rates, demonstrating that the proposed method outperforms the baselines.
\Fref{fig:confusion_matrix} illustrates the responses and their associated probabilities of our method tested on industrial connectors. While the top-1 accuracy is $16/19$, the top-3 accuracy reaches $100\%$, indicating that if the robot is allowed multiple attempts, it eventually succeeds in identifying the correct hole. Additionally, multiple \texttt{Yes} responses appear in incorrect pairs (non-diagonal cells), emphasizing the importance of leveraging generation probabilities to determine the most suitable hole.

\begin{figure}[t]
    \centering
    \includegraphics[width=0.8\linewidth]{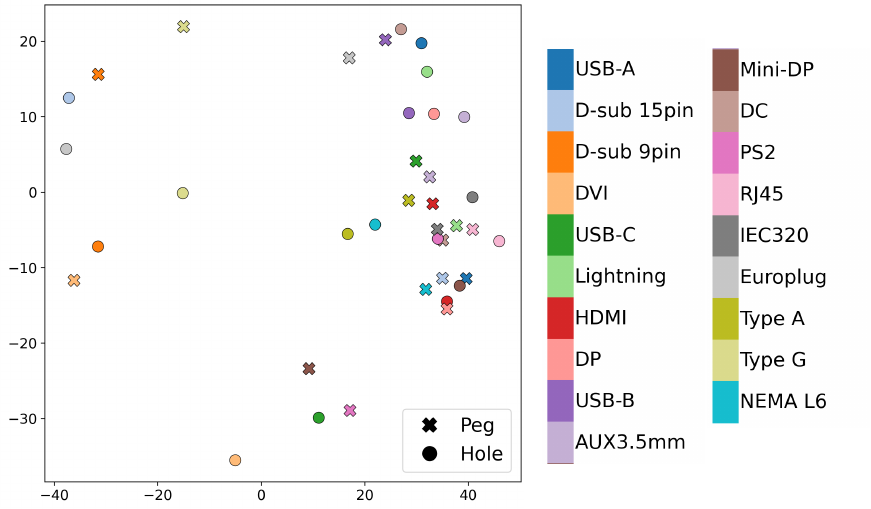}
    \caption{Visualization of t-SNE~\cite{vandermaaten2008tsne} plots of peg-hole pairs of industrial connectors. The inconsistent distances between correct pairs, i.e., the circle and cross with the same color, suggest simply measuring distance in latent space cannot determine peg-hole correspondences.}
    \label{fig:tsne}
\end{figure}

To gain deeper insights into the performance of \baselinefeature, we visualize the image feature vectors of pegs and holes using t-SNE~\cite{vandermaaten2008tsne} in \fref{fig:tsne}. The visualization reveals that corresponding pairs do not cluster closely, highlighting the difficulty of peg-hole matching relying on image features.

\begin{figure}[t]
    \centering
    \begin{subfigure}{0.45\linewidth}
        \includegraphics[width=\linewidth]{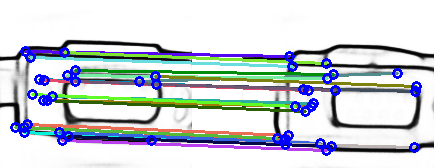}%
        \caption{Successful case}
    \end{subfigure}
    \begin{subfigure}{0.08\linewidth}
    \end{subfigure}
    \begin{subfigure}{0.45\linewidth}
        \includegraphics[width=\linewidth]{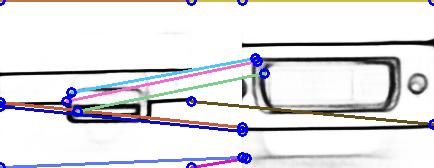}%
        \caption{Failed case}
    \end{subfigure}
    \caption{A successful and failed examples of \baselineomni\ baseline that uses edge detection~\cite{Pu2022EDTER} and feature matching~\cite{jiang2024Omniglue}. For each image, the left shows the edge image of a peg, and the right shows a hole. In the successful case (a), matching occurs only around the central connector area, whereas in the failed case (b), matching also happens in unrelated background regions, leading to an inaccurate estimation.}
    \label{fig:Omniglue}
    \vspace{-5mm}
\end{figure}

Furthermore, \fref{fig:Omniglue} presents the feature-matching results obtained using Omniglue on extracted edges. While edge detection combined with feature matching may seem like a straightforward approach---one frequently employed in prior works---it struggles in complex scenarios. The presence of intricate visual features in real-world images leads to failures in both edge detection and feature matching, further demonstrating the limitations of traditional techniques.

\subsection{Ablations on Different Inputs/Outputs} \label{subsec:ablation}
To further analyze the role of input representation and output processing in peg-hole matching, we conduct a series of ablation experiments by systematically modifying the input and output of the VLM and evaluating their impact on accuracy compared to our approach.

\textbf{Settings.}
We first investigate the impact of using multiple images from different angles. Specifically, we compare our method against two baselines: \emph{Cross-Sectional Image} and \emph{Angled Image}, where the input consists solely of either cross-sectional or angled images of the peg and hole. Additionally, \emph{3 Types of Image} includes a third viewpoint, providing a more comprehensive visual representation.

Next, we investigate alternative strategies for extracting information from the VLM while keeping the input images fixed. In \baselinename, we modify the prompt to generate names of the peg and hole (e.g., HDMI, USB-C) and determine a match by verifying whether the predicted names correspond.
In \emph{No Probability}, we constrain the output to a binary \texttt{Yes} or \texttt{No} response, discarding probability values and eliminating the ranking procedure described in Sec.~\ref{subsec:output_processing}. A hole is considered a match if it is the only one receiving a \texttt{Yes} prediction while all others are classified as \texttt{No}.

\textbf{Results.}
Table~\ref{tab:results_ablation} presents the success rates for various VLM use cases, showing that our method consistently outperforms all alternatives.
Both the \emph{Cross-Sectional Image} and \emph{Angled Image} cases drop their accuracies compared to ours, showing the importance of using multiple views for identifying a wide range of peg-hole pairs. Surprisingly, incorporating a third viewpoint in the \emph{3 Types of Image} case does not yield improvements. Instead, accuracy drops for toy puzzles and industrial connectors, possibly due to redundant or conflicting visual information confusing the VLM.

In \baselinename, predicting peg and hole names does not always ensure correct matches. For instance, in toy puzzles, a hole with a truck illustration was correctly labeled as truck, but its matching peg was mislabeled as car, causing a mismatch. Similarly, rare connectors were often misclassified with generic labels like power cable or electrical plug in industrial connectors, making identification difficult.
The \emph{No Probability} setting fails mostly due to multiple holes receiving a \texttt{Yes} response, causing ambiguity.

\begin{table}[t]
    \centering
    \caption{Success rate when using VLM in various ways.}
    \begin{tabular}{r|ccc} \toprule
         & 3D print & Toy puzzle & Connectors  \\ \midrule
         \emph{Cross-sectional Image} & \cellcolor{yellow!50}7/8 & 12/14 & 9/19 \\
         \emph{Angled Image} & 5/8 & 9/14  & 11/19\\
         \emph{3 Types of Image} & \cellcolor{yellow!50}7/8 & 12/14 & 15/19 \\
         \baselinename & 2/8 & 7/14 & 2/19 \\ 
         \emph{No Probability} & 0/8 & 3/14 & 0/19 \\
         Ours & \cellcolor{yellow!50}7/8 & \cellcolor{yellow!50}14/14 & \cellcolor{yellow!50}16/19 \\ 
         \bottomrule
    \end{tabular}
    \vspace{-3mm}
    \label{tab:results_ablation}
\end{table}

\subsection{End-to-End Evaluation}
Finally, we evaluate our method in a realistic scenario. While the evaluations in Sec.~\ref{subsec:exp_perception} and~\ref{subsec:ablation} assess performance in a controlled setting---where peg and hole images are captured in known poses with only a single hole per image---we now test its robustness in a more complex, real-world environment, with the full pipeline including hole detection, identification, SE(2) pose estimation, and insertion.

\begin{figure}[t]
    \centering
    \includegraphics[width=0.7\linewidth]{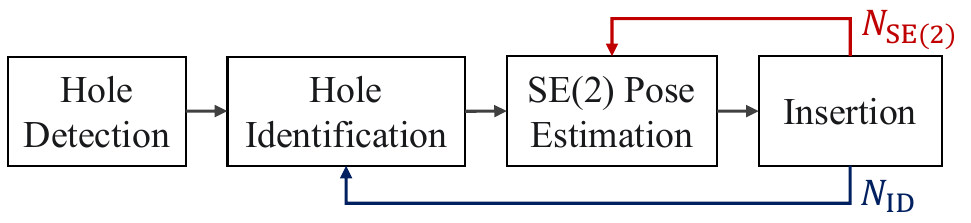}
    \caption{System level overview of our system. Our closed-loop insertion pipeline allows $N_\text{SE(2)}$ and $N_\text{ID}$ attempts to correct SE(2) pose and hole identification errors.}
    \label{fig:closed_loop}
    \vspace{2mm}
    \includegraphics[width=0.9\linewidth]{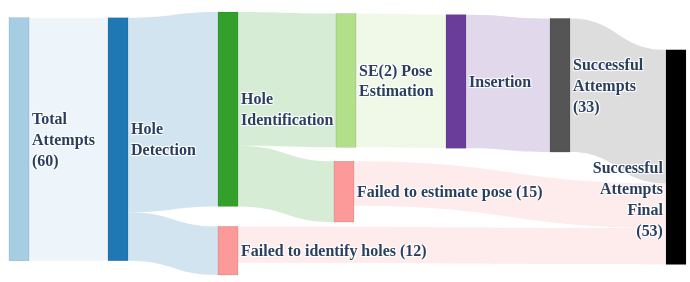}
    \caption{Analysis of how our method fails to achieve the task.}
    \label{fig:results_e2e}
    \vspace{2mm}
    \includegraphics[width=\linewidth]{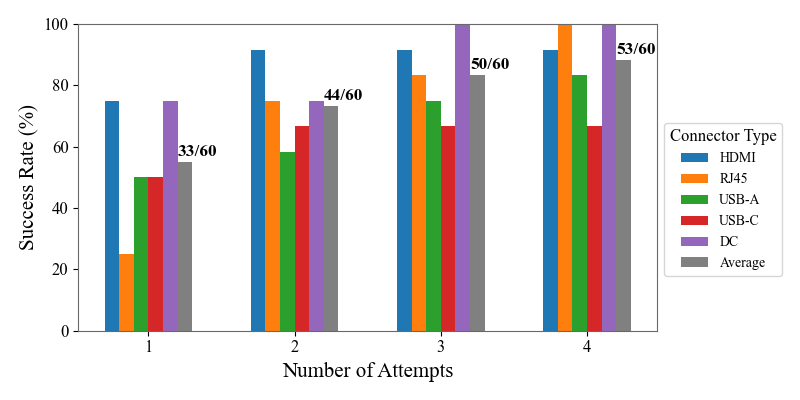}
    \vspace{-5mm}
    \caption{Success rate with multiple attempts for each connector type.}
    \label{fig:results_e2e_multi}
    \vspace{-5mm}
\end{figure}

\textbf{Settings.}
We conducted experiments using an Intel NUC mini PC, equipped with five distinct ports on its back panel: HDMI, USB-A, USB-C, RJ45, and DC input jack (see \fref{fig:intro_fig}). The NUC was randomly oriented in 12 different positions on a table, and we evaluated the overall end-to-end insertion success rate across all peg types, performing a total of 60 attempts. For the spiral search, we allow the robot to systematically explore within a maximum 5 mm radius following the Archimedes spiral search method, executing up to 10 rotations~\cite{jiang2022the}.
To enhance robustness, we implement a closed-loop system that enables multiple insertion attempts when failures occur. By explicitly estimating the rank and possible matches, our method allows the robot to retry up to $N_\text{SE(2)}=2$ times for SE(2) pose estimation and $N_\text{ID}=2$ times for hole identification, effectively compensating for estimation errors. The schematic is shown in \fref{fig:closed_loop}.

\textbf{Metrics.}
Since the ground-truth hole type and SE(2) pose are unknown for each NUC placement, we assess performance by insertion success or failure.

\textbf{Results.}
Our method achieved an initial success rate of 55.0\% (33 out of 60 attempts) on the first insertion attempt. \Fref{fig:results_e2e} presents a detailed breakdown of initial failures encountered during the insertion experiments. The primary source of initial failures was ambiguous hole identification, particularly evident when visually similar connectors were positioned closely together. For instance, the USB-C connector was occasionally misclassified as USB-A due to partial visibility of the USB-C hole within the cropped USB-A image. Additionally, SE(2) pose estimation errors occurred, notably with the RJ45 connector, where the small latch feature was occasionally misidentified in an upside-down orientation (flipped 180 degrees).

Crucially, the introduction of our iterative refinement process, which leverages a confidence-based ranking strategy (see \fref{fig:pipeline}) and allows up to two additional attempts to rectify identification and pose estimation errors, significantly enhanced performance. This adaptive approach effectively improved the overall success rate to 88.3\% (53 out of 60 attempts), as shown in \fref{fig:results_e2e_multi}.

Overall, these evaluations confirm our method's robust zero-shot generalization. The synergy between VLM-driven perception, precise pose estimation, and adaptive correction mechanisms enables highly reliable peg-in-hole assembly, demonstrating strong practical applicability in real-world robotic insertion tasks.

    \clearpage
\section{Conclusion and Discussion}\label{sec:conclusion}
Achieving zero-shot peg insertion remains a fundamental challenge in autonomous robotic assembly, requiring precise perception, identification, and pose estimation of mating parts without task-specific training. In this work, we introduced a novel VLM-driven framework that successfully identifies peg-hole compatibility and estimates SE(2) poses across a diverse set of peg-hole pairs, including 3D-printed parts, toy puzzles, and industrial connectors. Our method demonstrated exceptional generalization capabilities, achieving a 90.2\% accuracy in hole identification and an 88.3\% success rate in real-world insertion tasks.

Unlike traditional learning-based or feature-matching methods, our approach leverages VLMs' strong semantic understanding to recognize mating parts across unseen objects. By framing peg-hole matching as a compatibility assessment task, we effectively harnessed pre-trained models for zero-shot reasoning, eliminating the need for task-specific fine-tuning. Additionally, we integrate a confidence-aware ranking mechanism to resolve ambiguities, making it adaptable to real-world scenarios with multiple candidate holes.

Despite these advances, challenges remain in fine-grained pose estimation and insertion under varying conditions. Future work will focus on developing a closed-loop refinement system that dynamically adjusts hole identification and pose estimation based on real-time insertion feedback. Moreover, we aim to extend our approach to more complex, high-precision assembly tasks involving multi-modal sensor fusion, integrating tactile feedback and force control to further enhance robustness.

    \bibliographystyle{IEEEtran}
    \bibliography{references}

\end{document}